\documentclass[11pt,a4paper]{article}
\usepackage[hyperref]{acl2021}
\usepackage{times}
\usepackage{latexsym}
\usepackage{enumitem}
\usepackage{listings}
\usepackage{balance}
\usepackage{tikz}
\usepackage{pgfplots}
\usepackage{subcaption}
\usetikzlibrary{pgfplots.groupplots}

\usepackage[hang,flushmargin]{footmisc}
\usepackage{microtype}
\usepackage{booktabs}
\usepackage{amsfonts}
\usepackage{amsmath}
\usepackage{hyperref}
\usepackage{multirow}
\usepackage{graphicx}
\usepackage{enumitem}

\newcommand{\ignore}[1]{}

\aclfinalcopy

\title{A Replication Study of Dense Passage Retriever}

\author{Xueguang Ma, Kai Sun, Ronak Pradeep, and Jimmy Lin \\ [1ex]
  David R. Cheriton School of Computer Science \\
  University of Waterloo
}

\date{}

\begin{document}

\maketitle

\begin{abstract}
Text retrieval using learned dense representations has recently emerged as a promising alternative to ``traditional'' text retrieval using sparse bag-of-words representations.
One recent work that has garnered much attention is the dense passage retriever (DPR) technique proposed by \citet{karpukhin-etal-2020-dense} for end-to-end open-domain question answering.
We present a replication study of this work, starting with model checkpoints provided by the authors, but otherwise from an independent implementation in our group's Pyserini IR toolkit and PyGaggle neural text ranking library.
Although our experimental results largely verify the claims of the original paper, we arrived at two important additional findings that contribute to a better understanding of DPR:\
First, it appears that the original authors under-report the effectiveness of the BM25 baseline and hence also dense--sparse hybrid retrieval results.
Second, by incorporating evidence from the retriever and an improved answer span scoring technique, we are able to improve end-to-end question answering effectiveness  using exactly the same models as in the original work.
\end{abstract}

\section{Introduction}
\label{sec:intro}

Replicability and reproducibility form the foundation of the scientific enterprise.
Through such studies, we as a community gain increased confidence about the veracity of previously published results.
These investigations are often under-valued, especially compared to work that proposes novel models, but they nevertheless make important contributions to advancing science.

This paper presents a replicability study of the dense passage retriever (DPR) technique proposed by \citet{karpukhin-etal-2020-dense} for end-to-end open-domain question answering (QA).
To be precise, we use the term replicability in the sense articulated by the ACM,\footnote{\href{https://www.acm.org/publications/policies/artifact-review-and-badging-current}{Artifact Review and Badging}} characterized as ``different team, different experimental setup''.
We are able to achieve comparable measurements (i.e., effectiveness on different test collections) based on an independently developed computational artifact (i.e., a different implementation).
Specifically, our experiments rely on model checkpoints shared by the original authors, but we have otherwise built an entirely different implementation (other than the evaluation scripts).

DPR is worthy of detailed study because it represents an important exemplar of text retrieval using learned dense representations, which has recently emerged as a promising alternative to ``traditional'' text retrieval using sparse bag-of-words representations~\cite{Zhan:2006.15498:2020,Xiong:2007.00808:2020,Hofstatter:2010.02666:2020,Lin_etal_arXiv2020_DenseRanking}.
Our experiments largely verify the claims of \citet{karpukhin-etal-2020-dense} regarding the effectiveness of their proposed techniques.
However, we arrived at two important additional findings, one of which is inconsistent with the original work, the other of which presents an enhancement:

\begin{enumerate}[leftmargin=*]

\item Focusing on retrieval, we found that the effectiveness of the sparse retrieval (BM25) baseline is higher than values reported by \citet{karpukhin-etal-2020-dense}.
Whereas they reported that dense--sparse hybrid results do not meaningfully improve over dense retrieval alone, we arrived at the opposite conclusion, where hybrid techniques yield statistically significant gains.
We are able to achieve on average a three-point improvement in top-20 accuracy over the best DPR results across five standard QA test collections. 

\item Focusing on end-to-end QA effectiveness, we explored different techniques for evidence combination to extract the final answer span.
Whereas the original DPR paper only used scores from the reader to identify the final answer span, we investigated combining retriever scores and further experimented with the answer span selection technique described by~\citet{Mao:2009.08553:2020}.
In our best condition, we were able to achieve statistically significant improvements of around three points on exact match scores over the original DPR implementation, using the same exact models.

\end{enumerate}

\noindent The main contribution of this work is the replication of DPR, where our experimental results add a number of important refinements to the original work.
Code associated with our retrieval experiments is packaged in our Pyserini IR toolkit\footnote{\url{http://pyserini.io/}}~\cite{Lin2021PyseriniAE} and code associated with our end-to-end QA experiments is part of our PyGaggle toolkit\footnote{\url{http://pygaggle.ai/}} for neural text ranking.

\section{Methods}

DPR~\cite{karpukhin-etal-2020-dense} adopts the retriever--reader design proposed by~\citet{ChenDanqi_etal_ACL2017} for the open-domain QA task.
Both the task formulation and the pipeline architecture for tackling the task date from the late 1990s~\cite{Voorhees_Tice_TREC8}, so this general approach has a long history that predates neural networks.
The open-source code associated with the paper is available on GitHub (which we refer to as ``the DPR repo''),\footnote{\url{https://github.com/facebookresearch/DPR}} but it does not appear to contain code and models necessary to reproduce all results reported in the paper (more detailed discussions below).

\subsection{Retriever}
\label{methods:retriever}

In the retrieval stage, given a corpus $\mathcal{C} = \{D_1, D_2, ..., D_m\}$, the task is to return for each query $q$ a list of $k$ most relevant documents (i.e., most likely to contain the answer) from $\mathcal{C}$, where $k<<|\mathcal{C}|$.
In the original DPR paper and also our replication study, the corpus refers to a version of English Wikipedia (dump from 2018-12-20), and the ``documents'' are non-overlapping 100-word splits from the articles.

To be clear, in most text ranking applications, the ``unit of indexing'' (and also retrieval) is usually referred to as a ``document'' $D_j$, although in this case it is a passage (i.e., a split) from Wikipedia.
For consistency with this parlance, we use ``document'' and ``passage'' interchangeably throughout this paper.
To add to the potential confusion, results of the retriever are also referred to as ``contexts'' that are fed to the reader.

Dense retrieval with DPR uses a query encoder and a passage encoder, which are both based on BERT.
Queries and passages are encoded as dense representation vectors as follows:
$$q^*=\textrm{BERT}_q(q), D_j^*=\textrm{BERT}_D(D_j)$$
where $q^*$ and $D_j^*$ are low dimensional vectors (768).
The relevance score of a passage to a query is computed by dot product:
$$\textrm{Sim}(q, D_j) = \langle q^*, D_j^* \rangle$$
\noindent Thus, the top $k$ retrieval problem can be recast as a nearest neighbor search problem in vector space.
Operationally, this is accomplished via Facebook's Faiss library~\cite{Johnson:1702.08734:2017}.

\citet{karpukhin-etal-2020-dense} also investigated hybrid retrieval, combining results from dense retrieval (DPR) and sparse retrieval (BM25) by computing the linear combination of their respective scores to rerank the union of the two initial retrieved sets:
$$\lambda \cdot \textrm{Sim}(q, D_j) + \textrm{BM25}(q,D_j),$$
\noindent where $\lambda=1.1$, an empirical value tuned on the development set. 
BM25 retrieval was performed using Lucene with parameters $b=0.4$ and $k_1 = 0.9$.
However, the DPR repo does not appear to contain code for reproducing the BM25 and  hybrid fusion results.

We attempted to replicate the retriever results reported in~\citet{karpukhin-etal-2020-dense} with Pyserini, an IR toolkit that our group has been developing since 2019~\cite{Lin2021PyseriniAE}.
The toolkit supports sparse retrieval (i.e., BM25) via integration with another toolkit called Anserini~\cite{Yang2017AnseriniET}, which is built on Lucene.
Like in the original DPR work, Pyserini supports dense retrieval via integration with Facebook's Faiss library.
Combining dense and sparse retrieval, our toolkit supports hybrid retrieval as well.

To be clear, we started with model checkpoint releases in the DPR repo and did {\it not} retrain the query and passage encoders from scratch.
Otherwise, our implementation does not share any code with the DPR repo, other than evaluation scripts to ensure that results are comparable.

Similar to the original work, we calculated hybrid retrieval scores by linear combination of dense and sparse scores, as follows:
$$\textrm{Sim}(q, D_j) + \alpha\cdot \textrm{BM25}(q,D_j).$$
Note that, contrary to the original work, we placed the $\alpha$ weight on the BM25 score because this yields a more natural way to answer the pertinent research question:
Given dense retrieval as a starting point, does adding BM25 as an additional relevance signal provide any value?
This question is answered by a setting of $\alpha=0$, which is equivalent to discarding BM25 results.

Finally, there are a few more details of exactly how to combine BM25 and DPR scores worth exploring.
As a baseline, we tried using the raw scores directly in the linear combination (exactly as above).
However, we noticed that the range of scores from DPR and BM25 can be quite different.
To potentially address this issue, we tried the following normalization technique:\
If a document from sparse retrieval is not in the dense retrieval results, we assign to it the the minimum dense retrieval score among the retrieved documents as its dense retrieval score, and vice versa for the sparse retrieval score.

To arrive at a final top-$k$ ranking, the original DPR paper generated top $k'$ results from DPR and top $k'$ results from BM25 (where $k' > k$), before considering the union of the two result sets and combining the scores to arrive at the final top $k$.
\citet{karpukhin-etal-2020-dense} set $k'=2000$, but after some preliminary experimentation, we decided to fix $k'=1000$ in our experiments since it is a more common setting in information retrieval experiments (for example, $k=1000$ is the default in most TREC evaluations).

\subsection{Reader}

As is standard in a retriever--reader design, the retriever in~\citet{karpukhin-etal-2020-dense} returns $k$ candidate passages (i.e., splits from Wikipedia) for each query $q$.
The reader extracts the final answer span from these candidate contexts, where each context $C_i$ is comprised of the Wikipedia article title $C_i^\textrm{title}$ and the content text $C_i^\textrm{text}$.

The reader in DPR uses BERT-base and takes as input each candidate context $C_{i}$ concatenated to the question $q$. 
Answer extraction is treated as a labeling task, and the reader identifies the answer by predicting the start and end tokens of the answer span in the contexts. 
To do so, the DPR reader adds a linear layer on top of BERT to predict the start logit and end logit for each token from the final hidden layer representations. 
The score of an answer span is calculated by adding the start logit of the first token and the end logit of the last token.
The reader returns the $m$ highest scoring answer spans.
In addition, the reader uses the learned representation of [CLS] to predict the overall relevance of the context to the question.

In more detail, the reader operates as follows:
$$r_{i}, \mathcal{S} = \textrm{Reader}(\textrm{[CLS]} \ q \  \textrm{[SEP]} \ C_{i}^{\textrm{title}} \textrm{[SEP]} \ C_{i}^{\textrm{text}} ) $$
where $r_{i}$ is the overall relevance score for context $C_i$, and $\mathcal{S}$ comprises $m$ potential (answer span, span score) pairs extracted from context $C_i$:
$$\{ (S_{i,1}, s_{i,1}), (S_{i,2}, s_{i,2}), \ldots (S_{i,m}, s_{i,m})\}.$$ 
In the original paper, the final answer span is the candidate with the maximum span score from the context with the highest relevance score.

We attempted to replicate exactly the DPR implementation of answer extraction using our open-source PyGaggle neural reranking library, which holds the code to many of our other search-related projects.
Once again, we began with reader checkpoints released in the DPR repo, but otherwise our implementation is completely independent (other than, again, the evaluation code).

In addition to the answer extraction algorithm above, we also implemented the normalized answer span scoring technique described by~\citet{Mao:2009.08553:2020}.
Each answer span in each candidate context $C_i$ is rescored by:
$$s_{i,j}' = \textrm{softmax}(\vec{r})_i \cdot \textrm{softmax}(\vec{s_{i}})_j$$
where $\vec{r}=\{r_1, \cdots, r_k\}$ is the set of relevance scores of all candidate contexts and $\vec{s_i} = \{s_{i,1}, \cdots, s_{i,m}\}$ is the set of all span scores within context $C_i$.
Duplicate answer spans across all contexts are scored by accumulating their individual scores. 
The answer span with the maximum final score is selected as the final prediction.

In summary, we compared two answer span scoring techniques in the reader:\  the ``original'' answer span scoring technique described by~\citet{karpukhin-etal-2020-dense}, and the span scoring technique described by~\citet{Mao:2009.08553:2020}.

\subsection{Final Evidence Fusion}
\label{sec:fusion}

In the original DPR paper, the final answer span is only selected based on scores from the reader.
In our replication attempt, we additionally tried to exploit scores from the retriever to improve answer span selection. 
Our intuition is that predictions from both the retriever and the reader should contribute to the final answer.
Concretely, instead of just using the relevance score $r_{i}$ from the reader to score contexts, we fuse $r_{i}$ with the retriever score $R_{i}$, calculated by:
$$\beta\cdot r_{i} + \gamma\cdot R_{i}$$
Depending on the retrieval method, $R_{i}$ can be the sparse retrieval score, the dense retrieval score, or the score after hybrid fusion.
This final fused score replaces $r_{i}$ as the relevance score for each context in the answer span scoring step.
For example, with fusion, the answer span scoring technique of~\citet{Mao:2009.08553:2020} becomes $\textrm{softmax}(\beta\cdot\vec{r}+\gamma\cdot\vec{R})_i \cdot \textrm{softmax}(\vec{s_{i}})_j$.

Thus, to summarize, we explored four settings in our end-to-end QA replication:\ the original DPR span scoring technique, with and without retriever score fusion, and the answer span scoring technique of~\citet{Mao:2009.08553:2020}, with and without retriever score fusion.

\section{Experimental Setup}
\label{sec:setup}

\paragraph{Models}
Our replication efforts began with model checkpoints provided in the DPR repo.
Unfortunately,~\citet{karpukhin-etal-2020-dense} did not appear to make available all models used in their experiments, and thus, to be precise, our experiments used the following models:

\begin{itemize}[leftmargin=*]

\item Retriever$_{\textrm{NQ}}$:\ DPR encoders trained using just the NQ dataset (for the retriever).

\item Retriever$_{\textrm{Multi}}$:\ DPR encoders trained using a combination of datasets (for the retriever).

\item \textrm{Reader$_{\textrm{NQ-Single}}$}:\ the DPR reader trained on NQ with negative passages from retrieval results by Retriever$_{\textrm{NQ}}$. 

\item \textrm{Reader$_{\textrm{TQA-Multi}}$}:\ the DPR reader trained on TriviaQA with negative passages from retrieval results by Retriever$_{\textrm{Multi}}$.

\end{itemize}

\paragraph{Datasets}

We evaluated retrieval effectiveness on five standard benchmark QA datasets (NQ, TriviaQA, WQ, CuratedTREC, SQuAD), exactly the same as~\citet{karpukhin-etal-2020-dense}. 
We used the \textrm{Retriever$_{\textrm{Multi}}$} model, which can be applied to all five datasets.
For end-to-end QA, we evaluated on NQ and TriviaQA with the available models.
More precisely, we used the \textrm{Reader$_{\textrm{NQ-Single}}$} model to process the retrieved contexts from Retriever$_{\textrm{NQ}}$ for NQ and used the \textrm{Reader$_{\textrm{TQA-Multi}}$} model to process the retrieved contexts from Retriever$_{\textrm{Multi}}$ for TriviaQA.

\paragraph{Metrics}
For retrieval, we measured effectiveness in terms of top-$k$ retrieval accuracy,  defined as the fraction of questions that have a correct answer span in the top-$k$ retrieved contexts at least once.
End-to-end QA effectiveness is measured in terms of the exact match (EM) metric, defined as the fraction of questions that have an extracted answer span exactly matching the ground truth answer.

Missing from the original DPR paper, we performed significance testing to assess the statistical significance of metric differences.
In all cases, we applied paired $t$-tests at $p<0.01$; the Bonferroni correction was applied to correct for multiple hypothesis testing as appropriate.  

\paragraph{Hyperparameters}
In the hybrid retrieval technique described in the DPR paper, the $\lambda$ weight for combining dense and sparse retrieval scores is fixed to $1.1$.
However, our implementation replaces $\lambda$ with $\alpha$ (see Section~\ref{methods:retriever}).
We tuned the $\alpha$ values on different datasets by optimizing top-$20$ retrieval accuracy:
For datasets where we can obtain exactly same train/dev/test splits as the original DPR paper (NQ and TriviaQA), we tuned the weight on the development set. 
For the remaining datasets, where splits are not available or the original DPR paper does not provide specific guidance, we tuned the weight on a subset of the training data.
We obtained the optimal weight by performing grid search in the range $[0, 2]$ with step size $0.05$.

Similarly, for final evidence fusion, we tuned $\beta$ (i.e., the weight for the relevance score) and $\gamma$ (i.e., the weight for retriever score) on the development set of NQ and TriviaQA using grid search.
For greater computational efficiency, we performed tuning in multiple passes, first with a coarser step size and then with a finer step size.
For the original DPR answer span scoring technique, we fixed $\beta$ to one and performed a two-step grid search on $\gamma$.
We started with step size $0.05$ and found the optimal $\gamma_{1}$.
Then, we used step size $0.01$ in the range $[\gamma_{\textrm{1}} - 0.04, \gamma_{\textrm{1}} + $0.04$]$ to find the optimal $\gamma$.

For the answer span scoring technique of~\citet{Mao:2009.08553:2020}, we defined $\delta = \frac{\gamma}{\beta}$ and performed a three-step grid search on $\beta$ and $\delta$ (i.e., the weight for the retriever score becomes $\gamma = \beta \cdot \delta$).
We started with step size $0.2$ for both $\beta$ and $\delta$ and found the optimal pair of values $\beta_{\textrm{1}}, \delta_{\textrm{1}}$. 
We then repeated this process with step size $0.05$ and then $0.01$ in a smaller range around the optimal $\beta_i$ and $\delta_i$ from the previous pass.

For final evidence fusion, we tuned the weight parameters together with the number of retrieval results ($k$) up to $500$ with a step size of $20$.
Optimal parameters were selected based on the exact highest match score.

\section{Results}
\label{sec:results}

\subsection{Retrieval}

Table~\ref{table:retriever} reports top-$k = \{20, 100\}$ retrieval accuracy from our replication attempt, compared to figures copied directly from the original DPR paper; here we focus on results from Retriever$_{\textrm{Multi}}$.
The hybrid retrieval results reported in the original DPR paper is denoted Hybrid$_{\textrm{orig}}$, which is not directly comparable to either of our two techniques:\ Hybrid$_{\textrm{norm}}$ (with minimum score normalization) or Hybrid (without such normalization).
We make the following observations:

\begin{table}[t]
    \centering\scalebox{0.85}{
    \begin{tabular}{lllll}
    \toprule
         & \multicolumn{2}{c}{Top-20} & \multicolumn{2}{c}{Top-100} \\
        {\bf Condition} & orig & repl & orig & repl \\
        
        \toprule
        {\bf NQ} \\
          DPR & 79.4 & 79.5 & 86.0 & 86.1 \\
          BM25 & 59.1 & 62.9$^\dagger$ & 73.7 & 78.3$^\dagger$ \\
          Hybrid$_{\textrm{orig}}$ ($\lambda=1.1$) & 78.0 & - & 83.9 & - \\
          Hybrid$_{\textrm{norm}}$ ($\alpha=1.30$) & - & 82.6$^\ddag$ & - & 88.6$^\ddag$ \\
          Hybrid ($\alpha=0.55$) & -  & 82.7$^\ddag$ & - & 88.1$^\ddag$ \\
        \midrule
        {\bf TriviaQA} \\
          DPR & 78.8 & 78.9 & 84.7 & 84.8 \\
          BM25 & 66.9 & 76.4$^\dagger$ & 76.7 & 83.2$^\dagger$ \\
          Hybrid$_{\textrm{orig}}$ ($\lambda=1.1$) & 79.9 & - & 84.4 & - \\
          Hybrid$_{\textrm{norm}}$ ($\alpha=0.95$) & - & 82.6$^\ddag$ & - & 86.5$^\ddag$ \\
          Hybrid ($\alpha=0.55$) & -  & 82.3$^\ddag$ & - & 86.1$^\ddag$ \\
        \midrule
        {\bf WQ} \\
          DPR & 75.0 & 75.0 & 82.9 & 83.0 \\
          BM25 & 55.0 & 62.4$^\dagger$ & 71.1 & 75.5$^\dagger$ \\
          Hybrid$_{\textrm{orig}}$ ($\lambda=1.1$) & 74.7 & - & 82.3 & - \\
          Hybrid$_{\textrm{norm}}$ ($\alpha=0.95$) & - & 77.1$^\ddag$ & - & 84.4$^\ddag$ \\
          Hybrid ($\alpha=0.3$) & -  & 77.5$^\ddag$ & - & 84.0$^\ddag$ \\
        \midrule
        {\bf CuratedTREC} \\
          DPR & 89.1 & 88.8 & 93.9 & 93.4 \\
          BM25 & 70.9 & 80.7$^\dagger$ & 84.1 & 89.9$^\dagger$ \\
          Hybrid$_{\textrm{orig}}$ ($\lambda=1.1$) & 88.5 & - & 94.1 & - \\
          Hybrid$_{\textrm{norm}}$ ($\alpha=1.05$) & - & 90.1 & - & 95.0$^\ddag$ \\
          Hybrid ($\alpha=0.7$) & -  & 89.6 & - & 94.6$^\ddag$ \\
        \midrule
        {\bf SQuAD} \\
          DPR & 51.6 & 52.0 & 67.6 & 67.7 \\
          BM25 & 68.8 & 71.1$^\dagger$ & 80.0 & 81.8$^\dagger$ \\
          Hybrid$_{\textrm{orig}}$ ($\lambda=1.1$) & 66.2 & - & 78.6 & - \\
          Hybrid$_{\textrm{norm}}$ ($\alpha=2.00$) & - & 75.1$^\ddag$ & - & 84.4$^\ddag$ \\
          Hybrid ($\alpha=28$) & -  & 75.0$^\ddag$ & - & 84.0$^\ddag$ \\
        \bottomrule
    \end{tabular}}
    \caption{Retrieval effectiveness comparing results from the original DPR paper (``orig'') and our replication attempt (``repl''). The symbol $^\dagger$ on a BM25 result indicates effectiveness that is significantly different from DPR. The symbol $^\ddag$ indicates that the hybrid technique is significantly better than BM25 (for SQuAD) or DPR (for all remaining collections).}
    \label{table:retriever}
\end{table}

First, our dense retrieval results are very close to those reported in~\citet{karpukhin-etal-2020-dense}.
We consider this a successful replication attempt and our efforts add veracity to the effectiveness of the DPR technique.
Yay!

Second, our Pyserini BM25 implementation outperforms the BM25 results reported in the original paper across all datasets.
Furthermore, the gap is larger for $k=20$.
On average, our results represent a nearly seven-point improvement in top-20 accuracy and a nearly five-point improvement for top-100 accuracy. 
Since~\citet{karpukhin-etal-2020-dense} have not made available their code for generating the BM25 results, we are unable to further diagnose these differences.

Nevertheless, the results do support the finding that dense retrieval using DPR is (generally) more effective than sparse retrieval.
We confirmed that the effectiveness differences between DPR and BM25 in our replication results are statistically significant.
In all datasets except for SQuAD, DPR outperforms BM25; this is consistent with the original paper.
We further confirmed that for SQuAD, DPR is significantly worse than BM25.
As~\citet{karpukhin-etal-2020-dense} noted, Retriever$_{\textrm{Multi}}$ was trained by combining training data from all datasets but excluding SQuAD; these poor results are expected, since SQuAD draws from a very small set of Wikipedia articles.

Third, the effectiveness of hybrid dense--sparse fusion appears to be understated in the original DPR paper.
\citet{karpukhin-etal-2020-dense} found that hybrid retrieval is {\it less} effective than dense retrieval in most settings, which is inconsistent with our experimental results.
Instead, we found that dense--sparse retrieval consistently beats sparse retrieval across all settings.
The gains from both hybrid scoring techniques are statistically significant, with the exception of top-20 for CuratedTREC.
Our results might be due to better BM25 effectiveness, but we are unable to further diagnose these differences because, once again, the hybrid retrieval code is not provided in the DPR repo.
Further testing also found that the differences between the two hybrid techniques are not significant.
Thus, there does not appear to be a strong basis to prefer one hybrid technique over the other.

\begin{table}[]
\centering\scalebox{0.85}{
\begin{tabular}{lrrrr}
\toprule
{\bf Condition}    & $k=$ 20 & 100 & 500 & 1000 \\
\toprule
NQ          & 6.1  & 5.2   & 4.4    & 4.2    \\
TriviaQA    & 9.2  & 6.6   & 5.0    & 4.6    \\
WQ          & 5.9  & 5.9   & 5.8    & 5.7    \\
CuratedTrec & 6.9  & 7.2   & 6.3    & 5.9    \\
SQuAD       & 4.5  & 4.1   & 4.0    & 4.0    \\
\bottomrule
\end{tabular}}
\caption{The Jaccard overlap between sparse retrieval results and dense retrieval results.}
\label{table:overlap}
\end{table}

In Table~\ref{table:overlap}, we report overlap when taking different top-$k$ results from dense retrieval and sparse retrieval.
Overlap is measured in terms of Jaccard overlap, which is computed by the intersection over the union.
It is apparent that the overlap between dense and sparse results is quite small, which suggests that they are effective in very different ways.
This provides an explanation of why hybrid retrieval is effective, i.e., they are exploiting very different signals.
These results also justify the DPR design choice of retrieving $k' > k$ results from dense and sparse retrieval and then rescoring the union to arrive at the final top-$k$.

\subsection{End-to-End QA}

Table~\ref{table:e2e} presents results for our end-to-end question answering replication experiments on the NQ and TriviaQA datasets in terms of the exact match score.
The original results are shown in the ``orig'' column.
The ``repl'' column reports our attempt to replicate exactly the span scoring technique described in the original paper, whereas the ``GAR'' column shows results from using the technique proposed by~\citet{Mao:2009.08553:2020}.
The version of each technique that incorporates retriever scores (see Section~\ref{sec:fusion})~is denoted with a * symbol, i.e., ``repl*'' and ``GAR*''.
For NQ, we used Retriever$_{\textrm{NQ}}$ and Reader$_{\textrm{NQ-Single}}$; for TriviaQA, we used Retriever$_{\textrm{Multi}}$ and Reader$_{\textrm{TQA-Multi}}$.

\begin{table}[t]
    \centering\scalebox{0.85}{
    \begin{tabular}{llllll}
    \toprule
        {\bf Condition} & orig & repl & repl* & GAR & GAR* \\
        \toprule
        {\bf NQ} \\
        DPR & 41.5 & 41.2 & 42.5$^\dagger$ & 41.5 & 43.5$^\dagger$$^\ddag$ \\
        BM25 & 32.6 & 36.3 & 37.0 & 37.3$^\dagger$ & 38.4$^\dagger$$^\ddag$ \\
        Hybrid & 39.0 & 41.2 & 43.2$^\dagger$ & 41.9$^\dagger$ & 44.0$^\dagger$$^\ddag$ \\
        \midrule
        {\bf TriviaQA} \\
        DPR & 56.8 & 57.5 & 58.3$^\dagger$ & 58.9$^\dagger$ & 59.5$^\dagger$$^\ddag$ \\
        BM25 & 52.4 & 58.8 & 59.2 & 61.1$^\dagger$ & 61.6$^\dagger$$^\ddag$ \\
        Hybrid & 57.9 & 59.1 & 60.0$^\dagger$ & 61.0$^\dagger$ & 61.7$^\dagger$$^\ddag$ \\
        \bottomrule
    \end{tabular}}
    \caption{End-to-end QA effectiveness in terms of the exact match score, comparing different answer span scoring techniques. The ``orig'' and ``repl'' columns are the original and replicated results; ``GAR'' refers to the technique of~\citet{Mao:2009.08553:2020}; `*'' represents fusion of retriever scores. The symbol $^\dagger$ on a ``repl*'' result indicates stat sig.~improvement over ``repl''; on ``GAR'', over ``repl''; on ``GAR*'', over ``GAR''. The symbol $^\ddag$ on ``GAR*'' indicates sig.~improvement over ``repl''.}
    \label{table:e2e}
\end{table}

With retrieval using DPR only, the ``orig'' and ``repl'' scores on both datasets are close (within a point), which suggests that we have successfully replicated the results reported in~\citet{karpukhin-etal-2020-dense}.
Again, yay!

With retrieval using BM25 only, our replicated results are quite a bit higher than the original DPR results; this is not a surprise given that our BM25 results are also better.
When combining DPR and BM25 results at the retriever stage, the end-to-end effectiveness remains unchanged for NQ, but we observe a modest gain for TriviaQA.
The gain for TriviaQA is statistically significant.
So, it is {\it not} the case that better top-$k$ retrieval leads to improvements in end-to-end effectiveness.

\begin{figure*}[t]
\centering
\begin{tikzpicture}[scale = 0.6]
\begin{axis}[
width=0.75\textwidth,
height=0.60\textwidth,
legend cell align=left,
mark options={mark size=3},
axis y line*=left,
log ticks with fixed point,
every axis plot/.append style={ultra thick},
xmin=0, xmax=500,
ymin=30, ymax=45,
xtick={0, 100, 200, 300, 400, 500},
ytick={30, 33, 36, 39, 42, 45},
legend pos=south east,
xmajorgrids=true,
ymajorgrids=true,
xlabel= number of retrieval results ($k$),
ylabel= exact match score]

\addplot[
  solid, mark=*, red, mark options={scale=0.8,solid},
  error bars/.cd, 
    y fixed,
    y dir=both, 
    y explicit
] table [x=x, y=y, col sep=comma] {
    x,    y
    20, 30.02770083102493
    40, 31.74515235457064
    60, 33.1578947368421
    80, 33.767313019390585
    100, 34.15512465373961
    120, 34.37673130193906
    140, 34.79224376731302
    160, 34.847645429362885
    180, 34.986149584487535
    200, 35.124653739612185
    220, 35.20775623268698
    240, 35.54016620498615
    260, 35.59556786703601
    280, 35.59556786703601
    300, 35.62326869806094
    320, 35.7617728531856
    340, 35.87257617728532
    360, 35.95567867036011
    380, 36.038781163434905
    400, 36.038781163434905
    420, 36.177285318559555
    440, 36.260387811634345
    460, 36.260387811634345
    480, 36.20498614958449
    500, 36.31578947368421
};
\addlegendentry{\scalebox{0.6}{\textrm{BM25-repl}}}
\addplot[
  dashed, mark=*, red, mark options={scale=0.8,solid},
  error bars/.cd, 
    y fixed,
    y dir=both, 
    y explicit
] table [x=x, y=y, col sep=comma] {
    x,    y
    20, 30.360110803324098
    40, 32.21606648199446
    60, 33.795013850415515
    80, 34.37673130193906
    100, 34.958448753462605
    120, 35.041551246537395
    140, 35.48476454293629
    160, 35.51246537396121
    180, 35.59556786703601
    200, 35.6786703601108
    220, 35.84487534626039
    240, 36.177285318559555
    260, 36.31578947368421
    280, 36.34349030470914
    300, 36.45429362880886
    320, 36.64819944598338
    340, 36.78670360110804
    360, 36.92520775623269
    380, 37.091412742382275
    400, 37.119113573407205
    420, 37.202216066481995
    440, 37.257617728531855
    460, 37.257617728531855
    480, 37.229916897506925
    500, 37.285318559556785
};
\addlegendentry{\scalebox{0.6}{\textrm{BM25-GAR}}}
\addplot[
  dashed, mark=triangle*, red, mark options={scale=0.8,solid},
  error bars/.cd, 
    y fixed,
    y dir=both, 
    y explicit
] table [x=x, y=y, col sep=comma] {
    x,    y
    20, 30.80332409972299
    40, 33.04709141274238
    60, 34.70914127423823
    80, 35.59556786703601
    100, 36.066481994459835
    120, 36.50969529085872
    140, 36.84210526315789
    160, 36.89750692520776
    180, 37.202216066481995
    200, 37.257617728531855
    220, 37.285318559556785
    240, 37.61772853185595
    260, 37.56232686980609
    280, 37.67313019390582
    300, 37.75623268698061
    320, 37.92243767313019
    340, 38.03324099722992
    360, 38.03324099722992
    380, 38.06094182825485
    400, 38.088642659279785
    420, 38.17174515235457
    440, 38.282548476454295
    460, 38.282548476454295
    480, 38.227146814404435
    500, 38.365650969529085
};
\addlegendentry{\scalebox{0.6}{\textrm{BM25-GAR*}}}
\addplot[
  solid, mark=*, blue, mark options={scale=0.8,solid},
  error bars/.cd, 
    y fixed,
    y dir=both, 
    y explicit
] table [x=x, y=y, col sep=comma] {
    x,    y
    20, 40.83102493074792
    40, 41.19113573407202
    60, 41.10803324099723
    80, 41.16343490304709
    100, 41.0803324099723
    120, 41.0803324099723
    140, 40.96952908587257
    160, 40.80332409972299
    180, 40.85872576177285
    200, 40.7202216066482
    220, 40.470914127423825
    240, 40.3601108033241
    260, 40.30470914127424
    280, 40.27700831024931
    300, 40.08310249307479
    320, 40.0
    340, 39.94459833795014
    360, 39.86149584487534
    380, 39.80609418282548
    400, 39.83379501385041
    420, 39.83379501385041
    440, 39.80609418282548
    460, 39.91689750692521
    480, 39.83379501385041
    500, 39.77839335180056
};
\addlegendentry{\scalebox{0.6}{{\textrm{DPR-repl}}}}
\addplot[
  dashed, mark=*, blue, mark options={scale=0.8,solid},
  error bars/.cd, 
    y fixed,
    y dir=both, 
    y explicit
] table [x=x, y=y, col sep=comma] {
    x,    y
    20, 41.19113573407202
    40, 41.468144044321335
    60, 41.49584487534626
    80, 41.52354570637119
    100, 41.49584487534626
    120, 41.468144044321335
    140, 41.32963988919667
    160, 41.35734072022161
    180, 41.35734072022161
    200, 41.16343490304709
    220, 40.88642659279779
    240, 40.85872576177285
    260, 40.83102493074792
    280, 40.77562326869806
    300, 40.609418282548475
    320, 40.470914127423825
    340, 40.38781163434903
    360, 40.38781163434903
    380, 40.332409972299175
    400, 40.27700831024931
    420, 40.27700831024931
    440, 40.27700831024931
    460, 40.38781163434903
    480, 40.38781163434903
    500, 40.38781163434903
};
\addlegendentry{\scalebox{0.6}{\textrm{DPR-GAR}}}
\addplot[
  dashed, mark=triangle*, blue, mark options={scale=0.7,solid},
  error bars/.cd, 
    y fixed,
    y dir=both, 
    y explicit
] table [x=x, y=y, col sep=comma] {
    x,    y
    20, 42.32686980609418
    40, 42.90858725761773
    60, 42.853185595567865
    80, 43.01939058171745
    100, 43.07479224376731
    120, 43.18559556786703
    140, 43.2409972299169
    160, 43.35180055401662
    180, 43.46260387811635
    200, 43.40720221606648
    220, 43.37950138504155
    240, 43.35180055401662
    260, 43.37950138504155
    280, 43.37950138504155
    300, 43.32409972299169
    320, 43.32409972299169
    340, 43.37950138504155
    360, 43.40720221606648
    380, 43.35180055401662
    400, 43.32409972299169
    420, 43.35180055401662
    440, 43.37950138504155
    460, 43.40720221606648
    480, 43.37950138504155
    500, 43.37950138504155
};
\addlegendentry{\scalebox{0.6}{\textrm{DPR-GAR*}}}
\addplot[
  dashed, mark=triangle*, purple, mark options={scale=0.7,solid},
  error bars/.cd, 
    y fixed,
    y dir=both, 
    y explicit
] table [x=x, y=y, col sep=comma] {
    x,    y
    20, 43.29639889196676
    40, 43.46260387811635
    60, 43.49030470914128
    80, 43.822714681440445
    100, 43.988919667590025
    120, 43.905817174515235
    140, 43.933518005540165
    160, 43.905817174515235
    180, 44.01662049861496
    200, 43.905817174515235
    220, 43.878116343490305
    240, 43.905817174515235
    260, 43.933518005540165
    280, 43.933518005540165
    300, 43.961218836565095
    320, 43.878116343490305
    340, 43.878116343490305
    360, 43.850415512465375
    380, 43.850415512465375
    400, 43.76731301939058
    420, 43.76731301939058
    440, 43.73961218836565
    460, 43.795013850415515
    480, 43.73961218836565
    500, 43.73961218836565
};
\addlegendentry{\scalebox{0.6}{\textrm{Hybrid-GAR*}}}
\end{axis}
\end{tikzpicture}
\begin{tikzpicture}[scale = 0.6]
\begin{axis}[
width=0.8\textwidth,
height=0.60\textwidth,
legend cell align=left,
mark options={mark size=3},
axis y line*=left,
log ticks with fixed point,
every axis plot/.append style={ultra thick},
xmin=0, xmax=500,
ymin=53, ymax=63,
xtick={0, 100, 200, 300, 400, 500},
ytick={53, 55, 57, 59, 61, 63},
legend pos=south east,
xmajorgrids=true,
ymajorgrids=true,
xlabel= number of retrieval results ($k$),
ylabel= exact match score]

\addplot[
  solid, mark=*, red, mark options={scale=0.8,solid},
  error bars/.cd, 
    y fixed,
    y dir=both, 
    y explicit
] table [x=x, y=y, col sep=comma] {
    x,    y
    20, 55.78537965172809
    40, 57.02289401573412
    60, 57.385308936621584
    80, 57.68584813930876
    100, 57.862635905595326
    120, 58.05710244851057
    140, 58.3134447096261
    160, 58.30460532131176
    180, 58.41951736939804
    200, 58.410677981083715
    220, 58.50791125254132
    240, 58.58746574737028
    260, 58.64934146557058
    280, 58.6846990188279
    300, 58.72005657208521
    320, 58.72005657208521
    340, 58.72889596039954
    360, 58.711217183770884
    380, 58.72005657208521
    400, 58.79077167859984
    420, 58.79077167859984
    440, 58.81728984354283
    460, 58.82612923185716
    480, 58.84380800848581
    500, 58.83496862017148
};
\addlegendentry{\scalebox{0.6}{\textrm{BM25-repl}}}
\addplot[
  dashed, mark=*, red, mark options={scale=0.8,solid},
  error bars/.cd, 
    y fixed,
    y dir=both, 
    y explicit
] table [x=x, y=y, col sep=comma] {
    x,    y
    20, 57.09360912224874
    40, 58.64050207725625
    60, 59.091310881287015
    80, 59.4890833554318
    100, 59.754265004861665
    120, 59.96641032440555
    140, 60.187395032263765
    160, 60.3199858569787
    180, 60.39070096349333
    200, 60.381861575178995
    220, 60.523291788208255
    240, 60.655882612923186
    260, 60.77963404932378
    280, 60.86802793246707
    300, 60.91222487403871
    320, 60.8768673207814
    340, 60.8768673207814
    360, 60.8149916025811
    380, 60.8768673207814
    400, 60.947582427296034
    420, 60.974100592239026
    440, 61.00945814549633
    460, 61.035976310439324
    480, 61.08017325201096
    500, 61.11553080526827
};
\addlegendentry{\scalebox{0.6}{\textrm{BM25-GAR}}}
\addplot[
  dashed, mark=triangle*, red, mark options={scale=0.8,solid},
  error bars/.cd, 
    y fixed,
    y dir=both, 
    y explicit
] table [x=x, y=y, col sep=comma] {
    x,    y
    20, 57.385308936621584
    40, 58.95872005657209
    60, 59.5509590736321
    80, 59.95757093609122
    100, 60.293467692035705
    120, 60.558649341465575
    140, 60.708918942809156
    160, 60.8149916025811
    180, 60.84150976752408
    200, 60.885706709095736
    220, 60.95642181561036
    240, 61.1243701935826
    260, 61.22160346504022
    280, 61.29231857155485
    300, 61.36303367806948
    320, 61.35419428975515
    340, 61.39839123132679
    360, 61.35419428975515
    380, 61.460266949527096
    400, 61.46910633784142
    420, 61.44258817289844
    440, 61.50446389109874
    460, 61.530982056041715
    480, 61.566339609299035
    500, 61.53982144435604
};
\addlegendentry{\scalebox{0.6}{\textrm{BM25-GAR*}}}
\addplot[
  solid, mark=*, blue, mark options={scale=0.8,solid},
  error bars/.cd, 
    y fixed,
    y dir=both, 
    y explicit
] table [x=x, y=y, col sep=comma] {
    x,    y
    20, 55.02519225669583
    40, 55.953328029700344
    60, 56.39529744541678
    80, 56.607442764960666
    100, 56.801909307875896
    120, 56.75771236630425
    140, 56.84610624944754
    160, 56.81074869619023
    180, 56.890303191019186
    200, 56.93450013259083
    220, 56.97869707416247
    240, 57.04057279236277
    260, 57.084769733934415
    280, 57.102448510563065
    300, 57.11128789887739
    320, 57.137806063820385
    340, 57.137806063820385
    360, 57.182003005392026
    380, 57.270396888535316
    400, 57.2969150534783
    420, 57.385308936621584
    440, 57.420666489878904
    460, 57.44718465482188
    480, 57.509060373022194
    500, 57.5444179262795
};
\addlegendentry{\scalebox{0.6}{\textrm{DPR-repl}}}
\addplot[
  dashed, mark=*, blue, mark options={scale=0.8,solid},
  error bars/.cd, 
    y fixed,
    y dir=both, 
    y explicit
] table [x=x, y=y, col sep=comma] {
    x,    y
    20, 56.138955184301246
    40, 57.15548484044904
    60, 57.5444179262795
    80, 57.96870856536728
    100, 58.19853266153983
    120, 58.25156899142579
    140, 58.33996287456908
    160, 58.331123486254754
    180, 58.52559002916998
    200, 58.63166268894192
    220, 58.82612923185716
    240, 58.78193229028551
    260, 58.8703261734288
    280, 58.83496862017148
    300, 58.8703261734288
    320, 58.79077167859984
    340, 58.77309290197118
    360, 58.77309290197118
    380, 58.76425351365685
    400, 58.79961106691417
    420, 58.86148678511447
    440, 58.91452311500044
    460, 58.89684433837179
    480, 58.90568372668611
    500, 58.86148678511447
};
\addlegendentry{\scalebox{0.6}{\textrm{DPR-GAR}}}
\addplot[
  dashed, mark=triangle*, blue, mark options={scale=0.7,solid},
  error bars/.cd, 
    y fixed,
    y dir=both, 
    y explicit
] table [x=x, y=y, col sep=comma] {
    x,    y
    20, 56.174312737558566
    40, 57.235039335278
    60, 57.65932997436577
    80, 58.0659418368249
    100, 58.242729603111464
    120, 58.286926544683105
    140, 58.47255369928401
    160, 58.59630513568461
    180, 58.72889596039954
    200, 58.7465747370282
    220, 58.8703261734288
    240, 58.98523822151507
    260, 59.01175638645806
    280, 59.047113939715366
    300, 59.08247149297269
    320, 59.108989657915664
    340, 59.188544152744626
    360, 59.223901706001946
    380, 59.285777424202244
    400, 59.40068947228851
    420, 59.43604702554583
    440, 59.43604702554583
    460, 59.43604702554583
    480, 59.4890833554318
    500, 59.524440908689115
};
\addlegendentry{\scalebox{0.6}{\textrm{DPR-GAR*}}}
\addplot[
  dashed, mark=triangle*, purple, mark options={scale=0.7,solid},
  error bars/.cd, 
    y fixed,
    y dir=both, 
    y explicit
] table [x=x, y=y, col sep=comma] {
    x,    y
    20, 60.67356138955184
    40, 61.08901264032529
    60, 61.318836736497836
    80, 61.43374878458411
    100, 61.41607000795545
    120, 61.522142667727394
    140, 61.575178997613364
    160, 61.530982056041715
    180, 61.575178997613364
    200, 61.5840183859277
    220, 61.60169716255635
    240, 61.619375939185005
    260, 61.73428798727129
    280, 61.6812516573853
    300, 61.73428798727129
    320, 61.6812516573853
    340, 61.654733492442325
    360, 61.672412269070975
    380, 61.69009104569964
    400, 61.645894104128
    420, 61.59285777424203
    440, 61.61053655087068
    460, 61.62821532749933
    480, 61.672412269070975
    500, 61.645894104128
};
\addlegendentry{\scalebox{0.6}{\textrm{Hybrid-GAR*}}}
\end{axis}

\end{tikzpicture}
\caption{End-to-end question answering effectiveness (exact match score) varying the number of retrieval results ($k$) for NQ (left) and TriviaQA (right).}
\label{figure:e2e-k}
\end{figure*}
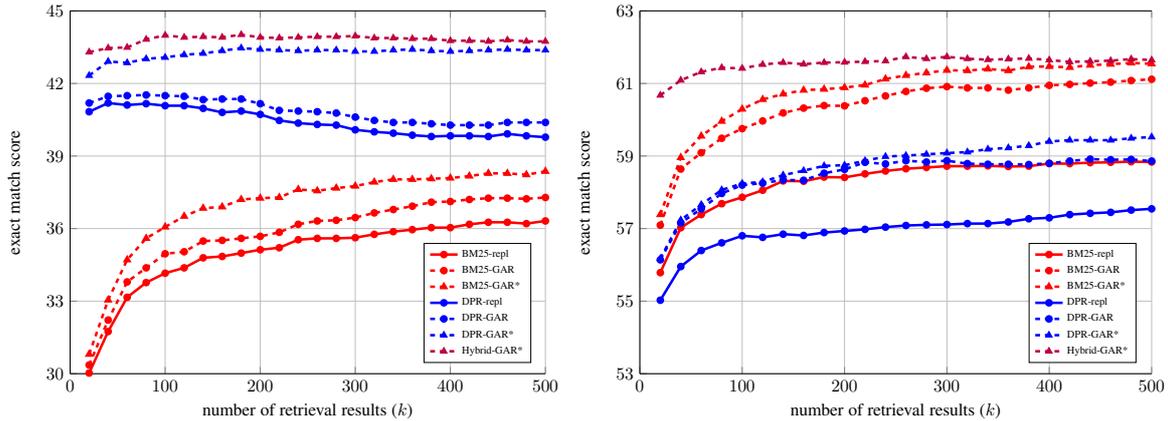

Comparing the ``repl'' and ``repl*'' columns, we observe that combining scores from the retriever yields modest gains across all conditions.
These gains are significant for four out of the six conditions, which suggests that retriever scores contribute to improving effectiveness.
Comparing the ``GAR'' and ``repl'' columns, we also observe modest gains when adopting the answer span selection technique of~\citet{Mao:2009.08553:2020}.
These gains are significant for all except one condition.
Comparing the ``GAR'' and ``GAR*'' columns, we find that in all cases, incorporating retriever scores significantly increases effectiveness.

Finally, putting everything together---using both the answer span scoring technique of~\citet{Mao:2009.08553:2020} and incorporating retriever scores---we observe statistically significant gains across all retrieval conditions, as can be seen in the ``GAR*'' vs.\ ``repl'' columns across all rows.
Compared to the best replicated results, we obtained an improvement of approximately three points in end-to-end QA effectiveness compared to the best answer extraction approach described in~\citet{karpukhin-etal-2020-dense}.
Note that we were able to obtain these improvements using exactly the model checkpoints provided in the DPR repo---we have simply added two relatively simple tricks to improve scoring and evidence combination.

In Figure~\ref{figure:e2e-k}, we plot exact match scores as a function of varying $k$ retrieval results for NQ (left) and TriviaQA (right).
That is, we show how end-to-end QA effectiveness changes as the reader is provided more contexts from the retriever to consider.
There are two factors here at play:\ On the one hand, top-$k$ accuracy increases monotonically, i.e., as $k$ increases, so does the likelihood that the answer appears in the contexts fed to the reader.
On the other hand, the reader is asked to consider more contexts, and thus needs to discriminate the correct answer from a larger pool of candidate contexts, some of which might be low quality and thus serve as ``distractors'' from the correct answer.
How do these factors balance out?
Similar analyses in previous work with BM25 retrieval have shown that end-to-end QA effectiveness increases with increasing $k$~\cite{Yang_etal_NAACL2019demo,Xie_etal_WWW2020}; that is, the reader does not appear to be ``confused'' by the non-relevant material.
Indeed, in our BM25 results we also observe the same trend.

Interestingly, however, when we switch from BM25 results to DPR results, the behavior appears to change.
For TriviaQA, the effectiveness curve behaves as expected, but for NQ, the exact match score trends up and then decreases after a peak.
This means that while the likelihood of the reader seeing a correct answer in the candidate contexts increases with $k$, it is more likely to be negatively affected by increasing amounts of non-relevant contexts as well.
This general behavior is also seen for the hybrid scoring techniques:\ as $k$ increases, so does the exact match score, but only up to a certain point.
Beyond this point, feeding the reader more candidate contexts leads to slight decreases in end-to-end effectiveness.

\section{Conclusion}

The breakneck pace at which NLP and IR are advancing, we argue, makes reproducibility and replicability critical to advancing science---to ensure that we are building on a firm foundation.
Our study adds to the veracity of the claims made by~\citet{karpukhin-etal-2020-dense}, and our work does indeed confirm that DPR is an effective dense retrieval technique.
However, we arrived at two important additional findings, one of which is inconsistent with the original work, the other of which presents an enhancement.
Together, they enrich our understanding of DPR.

\section{Acknowledgments}

This research was supported in part by the Canada First Research Excellence Fund and the Natural Sciences and Engineering Research Council (NSERC) of Canada.
Computational resources were provided by Compute Ontario and Compute Canada.

\bibliographystyle{acl_natbib}
\bibliography{main}

\end{document}